\documentclass[10pt,twocolumn,letterpaper]{article}

\usepackage{iccv}
\usepackage{times}
\usepackage{epsfig}
\usepackage{graphicx}
\usepackage{amsmath}
\usepackage{amssymb}
\usepackage{booktabs}
\usepackage{booktabs}
\usepackage{float}
\usepackage{multirow}
\usepackage{subfigure}

\usepackage[breaklinks=true,bookmarks=false]{hyperref}

\iccvfinalcopy % *** Uncomment this line for the final submission

\def\iccvPaperID{****} % *** Enter the ICCV Paper ID here

\ificcvfinal\fi

\begin{document}

%%%%%%%%% TITLE
\title{From Single to Multiple: Leveraging Multi-level Prediction Spaces \\
	for Video Forecasting}

\author{Mengcheng Lan$^{1^*}$ \qquad Shuliang Ning$^{12^*}$ \qquad Yanran Li$^1$ \qquad Qian Chen$^3$ \\ Xunlai Chen$^{3}$\footnotemark[2] \qquad Xiaoguang Han$^{12}$\footnotemark[2] \qquad Shuguang Cui$^{12}$ \\
$^1$Shenzhen Research Institute of Big Data \quad $^2$The Chinese University of Hong Kong (Shenzhen) \\ $^3$Moteorological Bureau of Shenzhen\\
{\tt\small lanmengchengds@gmail.com \ shuliangning@link.cuhk.edu.cn \ liy@bournemouth.ac.uk} \\ {\tt\small \{chenqian,chenxunlai\}@weather.sz.gov.cn \ \{hanxiaoguang,shuguangcui\}@cuhk.edu.cn}
}

\maketitle

\renewcommand{\thefootnote}{\fnsymbol{footnote}}
\footnotetext[1]{These authors contributed equally to this work.} 
\footnotetext[2]{Corresponding authors.}

\ificcvfinal\thispagestyle{empty}\fi

%%%%%%%%% ABSTRACT
\begin{abstract}
	Despite video forecasting has been a widely explored topic in recent years, the mainstream of the existing work still limits their models with a single prediction space but completely neglects the way to leverage their model with multi-prediction spaces. This work fills this gap. For the first time, we deeply study numerous strategies to perform video forecasting in multi-prediction spaces and fuse their results together to boost performance. The prediction in the pixel space usually lacks the ability to preserve the semantic and structure content of the video however the prediction in the high-level feature space is prone to generate errors in the reduction and recovering process. Therefore, we build a recurrent connection between different feature spaces and incorporate their generations in the upsampling process. Rather surprisingly, this simple idea yields a much more significant performance boost than PhyDNet (performance improved by 32.1\% MAE on MNIST-2 dataset, and 21.4\% MAE on KTH dataset). Both qualitative and quantitative evaluations on four datasets demonstrate the generalization ability and effectiveness of our approach. We show that our model significantly reduces the troublesome distortions and blurry artifacts and brings remarkable improvements to the accuracy in long term video prediction.  The code will be released soon.  
\end{abstract}

\begin{figure}[t]
	\begin{center}
		\includegraphics[width=1\linewidth]{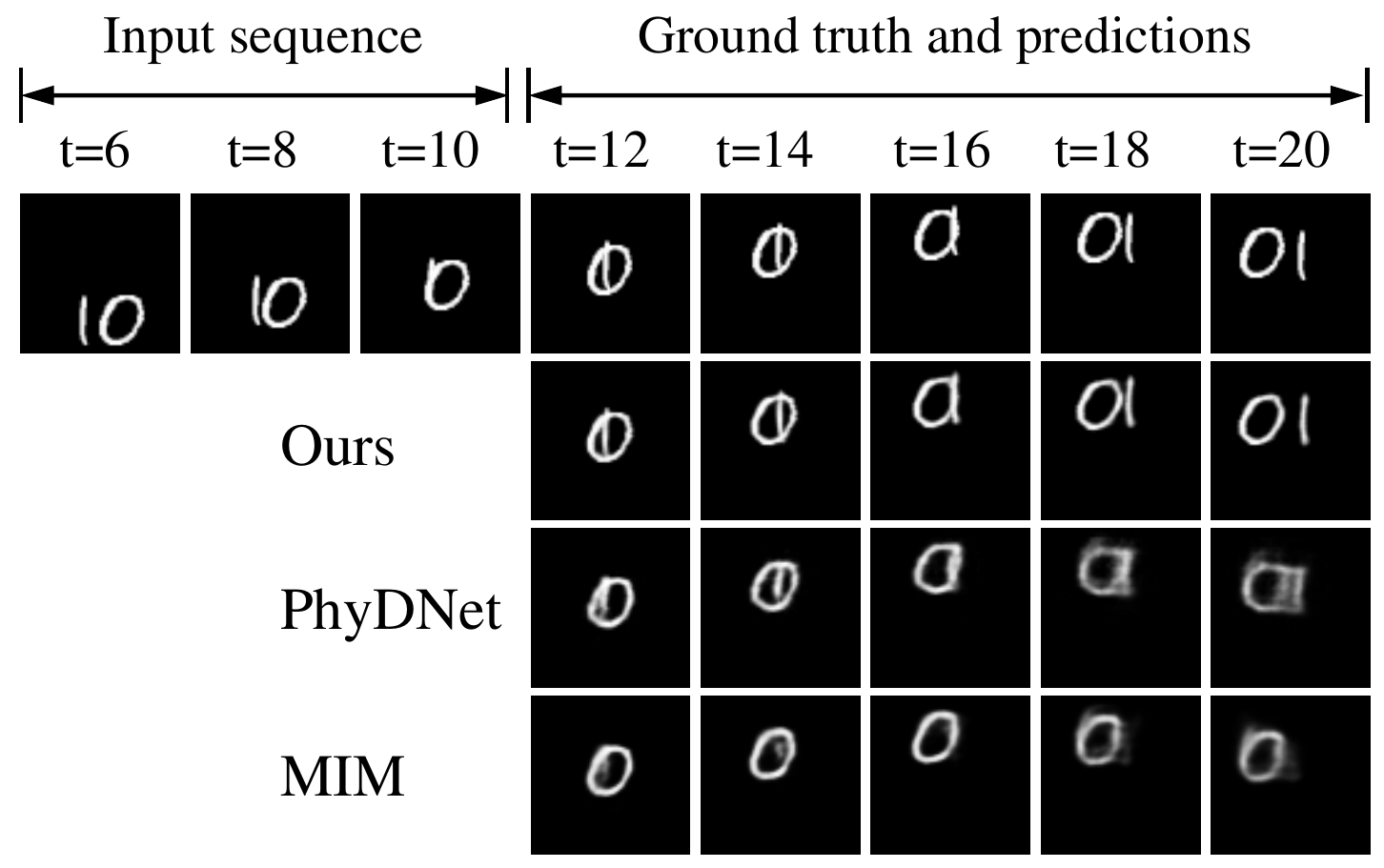}
	\end{center}
	\caption{\textbf{Comparison results of our model with the state-of-the-art on the moving MNIST dataset \cite{srivastava2015unsupervised}. } The MIM \cite{wang2019memory} model predicts the future frames on Pixel Space and the PhyDNet \cite{guen2020disentangling} model conducts the prediction on high-level feature space. }
	\label{fig:Teaser}
\end{figure}

%%%%%%%%% BODY TEXT
\section{Introduction}
Video prediction is a crucial problem in Computer Vision that aims to anticipate the future frames conditioned on an observed video clip. It is a challenging task, especially when it comes to long term prediction due to the drastic evolution of the pixel space and the complicated uncertainty of the future. Tremendous accomplishments have been made through the introduction of Long Short-Term Memory (LSTM) models and their variants to address this task, and in fact, the current trend is to mainly focus on designing sophisticated architectures to exploit the spatial-temporal clues. However, none of these existing approaches has paid particular attention to the prediction space used in the video prediction task.

In this paper, we first carry out an in-depth investigation of the existing work from the neglected prediction space perspective and reveal that a simple but effective strategy to utilize multi-prediction spaces can surprisingly bring about marginal gains over the recent leading work than excepted. 
%For instance, we show that our model based on four-prediction spaces brings as high as 21.0 MSE and 56.3 MAE improvements on the existing benchmark MNIST-3 \cite{yu2019efficient} in comparison with PhyDNet \cite{guen2020disentangling}.
For instance, we show that our model based on four-prediction spaces gains an improvement of $29.5\%$ and $32.1\%$ in terms of MSE (17.2) and MAE (47.7) respectively on the existing benchmark MNIST-2 in comparison with PhyDNet \cite{guen2020disentangling}.
We demonstrate an example of comparison results in Figure \ref{fig:Teaser}, from which we can observe that our approach can correctly predict the movement of two numbers.

The feature space used as the input and output of the recurrent prediction process is noted as prediction space. According to the prediction space used, the attempts of existing research can be broadly categorized into two types. The first type of approaches conducts prediction directly in the pixel space \cite{xingjian2015convolutional, wang2017predrnn, wang2018predrnn++, wang2019memory, wang2018eidetic, terwilliger2019recurrent,wu2021motionrnn}, where their recurrent models directly predict the evolution of pixels and recursively generate future frames. Despite their advantages in modelling the sequence data, these methods present two main limitations: first, these models are limited to a pixel-to-pixel level prediction, and as a result they have difficulties learning to hold the semantic and structural information in the videos, e.g. shape, motion; secondly, errors in recurrent models are prone to accumulate in such a way that small discrepancies at the beginning will be amplified to serious compound error and distorted structures in the long term. 

Another group of trending works tackles these limitations by predicting the video in high-level feature space first, then translating it back into pixel-level frames \cite{guen2020disentangling, villegas2017decomposing, villegas2018hierarchical, yang2018pose, denton2018stochastic}. As memtioned in \cite{oprea2020review}, researchers have proposed different frameworks to predict the instant segmentation \cite{chiu2020segmenting}, motion \cite{villegas2017decomposing}, human pose \cite{yang2018pose}, object trajectory \cite{wu2020future}, or key points \cite{kim2019unsupervised} in the video first, then translate them back into the pixel space. In this way, the video prediction task is simplified and the content in the video is better preserved, thus leading to a decrease of the pixel level error due to more accurate content. However, the reduction and recovering processes between the pixel-level space and the high-level feature space usually bring intractable errors. 

%The first type focuses on designing complex recurrent architectures to memorize the complicated variations in spatial and temporal space, whilst the second type of approaches dedicate to learn a proper high-level representation that preserves the macroscopical structure and content of the video in the long term prediction well. 

Although numerous techniques have been introduced so far, most of the existing work only limited the prediction task to one single feature space. Performing video prediction simultaneously in multi-prediction spaces has never been investigated in the literature to our best knowledge. We believe that the different level of feature spaces can be complementary to each other.

In light of that, our key insight is to conduct the prediction task at multi-feature spaces and recover the final frame by assembling all the prediction results. In this paper, we dedicatedly investigate the prediction performance in different feature spaces and propose an architecture with Recurrent Connections. Specifically, the input video is downsampled through convolutional layers to obtain different feature spaces. After that, an LSTM-based module is employed to predict the feature sequences on different scales. The final frames will be synthesised by fusing all the results together. 

%To fully digest the advantage of each feature space, we thus through analysis various ways to build up the recurrent connection and fuse the prediction results. Furthermore, the effectiveness of the same strategy is introduced on another semantic feature space to validate the generalization ability. 

Our main contributions can be summarized as follows:

\begin{enumerate}
	\item To our knowledge, we are the first to take notice of the previously unrealised multi-prediction space perspective and propose a novel recurrent connection to address the video prediction problem.
	\item The experiment results show that our model introduces considerable improvements compared to the state-of-the-art works on various challenging video datasets. Furthermore, the generalization of our idea is validated by introducing the same strategy on different architectures.
	\item We thoroughly analyze various ways to build the recurrent connection and fuse the prediction results together from multi-spaces. A series of comprehensive experiments are conducted to provide a variety of insights for the utilization of multi-prediction spaces. 
\end{enumerate}

%%%%%%%%% Related work 
\section{Related Works}  
We reviewed the current approaches to tackle the task of video prediction from the perspective of the prediction space being used. They usually adopt two types, the pixel space and high-level feature space. 

\noindent \textbf{Pixel Space.}
Many recent works \cite{lotter2016deep,xingjian2015convolutional,shi2017deep,wang2019memory,wu2021motionrnn} attempted to directly predict future pixel intensities without any explicit modeling of the scene dynamics. Ranzato et al. \cite{ranzato2014video} constructed a Recurrent Neural Network (RNN) model to predict the next frames. Inspired by architectures used for language processing, Srivastava et al. \cite{srivastava2015unsupervised} introduced the sequence-to-sequence model to video prediction. Since this kind of model can only capture temporal variations, to instead learn both spatial and temporal variations in a unified network structure, Shi et al. \cite{xingjian2015convolutional} integrated the convolutional operator into recurrent state transition functions and proposed the Convolutional LSTM for joint modelling of both variations. However, in their approach, the LSTM cells in different layers are independent of each other and consequently the information from the last top layer cannot flow into the bottom layer during the next time step. In order to address this issue, \cite{lotter2016deep,finn2016unsupervised,kalchbrenner2017video,shi2017deep,wang2017predrnn,wang2018predrnn++,wang2019memory} further extended the convolutional LSTM model and investigated spatio-temporal prediction. All of these models focus on rectifying the LSTM cells to capture more spatial and temporal clues.

Furthermore, more various strategies are carried out. Under the assumption that video sequences are symmetric, Kwon et al. \cite{kwon2019predicting} implemented a retrospective prediction network by training a generator for both, forward and backward predictions. Hu et al. \cite{hu2019novel} presented a novel cycle-consistency loss function to be incorporated within their framework for more accurate predictions. In addition, several other studies \cite{misra2016shuffle,hou2019video,oliu2018folded} further explored the benefits of forward and backward predictions. Another typical way of predicting video frames is to use 3D convolutions. Wang et al. \cite{wang2018eidetic} presented a gated-controlled self-attention module that effectively manages historical memory records across multiple time steps.  

Overall, the performance of these models still suffers from a significant degradation in the long-term prediction task. The high dimensionality of the pixel space causes the prediction error to grows exponentially as the prediction goes longer.

\noindent \textbf{High-Level Feature Space.}
%To deal with the curse of dimensionality, various approaches reduced the prediction space to high-level representations. 
To deal with the curse of dimensionality, researchers proposed various approaches to learn the high-level representations in the latent space of videos via an encode networks first, then perform prediction in the learned high-level space. 

%such as semantic and instance segmentation, keypoints representation and human pose. As a matter of fact, high-level feature space has recently attracted increasing interest and has emerged as a promising avenue to simplify the video prediction task, due to its ability to largely retain the semantics and structure better.

The proposed Parsing with predictive feature Learning (PEARL) PEARL \cite{jin2017video} is the first systematic predictive learning model for video scene parsing. Concurrently, Luc et al. \cite{luc2017predicting} extended the msCNN model of \cite{mathieu2015deep} to the novel task of predicting semantic segmentation of future frames. However, both these models are not end-to-end and do not explicitly capture the temporal continuity across frames. To address this limitation, Jin et al. \cite{jin2017predicting} ﬁrst proposed a model for jointly predicting motion ﬂow and scene parsing. An advantage of flow-based representations is that they implicitly draw temporal correlations from the input data, thus producing temporally coherent per-pixel segmentation. Aside from segmentation, other high-level spaces such as human pose and keypoints represent promising ways of approaching this problem. Minderer et al. \cite{minderer2019unsupervised} modelled dynamics in the keypoint coordinate space, achieved stable learning and avoided the compounding of errors in pixel space. Tang et al. \cite{tang2019pose} proposed a pose-guided approach for appearance preserving video prediction by combining global and local information using Generative Adversarial Networks (GANs). However, a major drawback of these semantic feature spaces is that they are not generic for all the video data and usually requires laborious ground-truth labelling.
Yu et al. \cite{yu2019crevnet} proposed the Conditionally Reversible Network (CrevNet) that theoretically ensures no information loss during the feature extraction process, in this way guaranteeing a higher level of efficiency for their model as well.

\begin{figure*}[t]
	\begin{center}
		\includegraphics[width=1\linewidth]{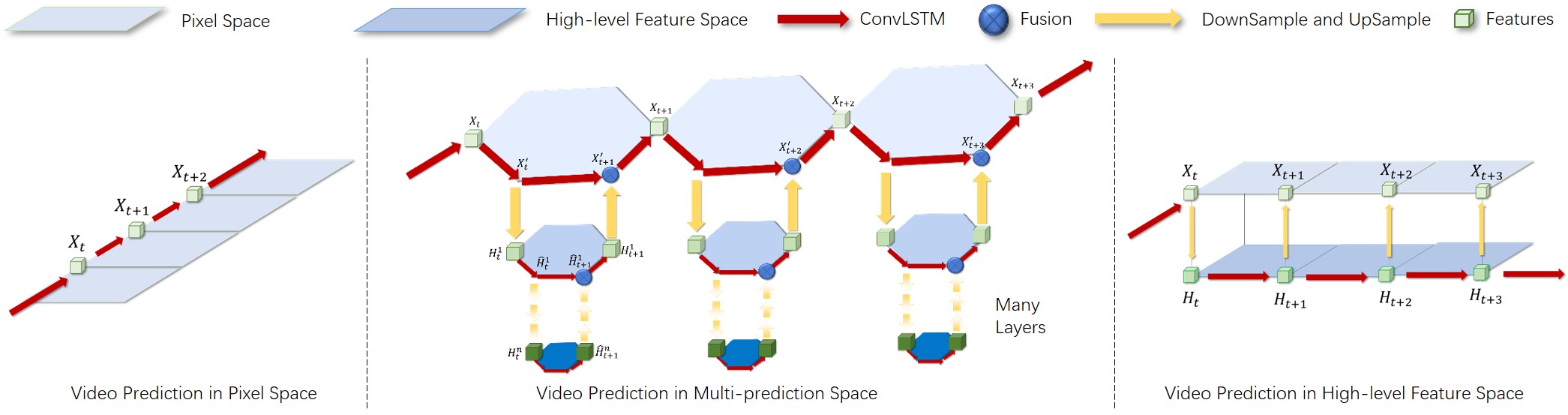}
	\end{center}
	\caption{\textbf{The workflow of the three types of approaches.} Our model is presented in the middle image. The network architecture will repeat the same structure of the first two layers when the prediction spaces go deeper.}
	\label{fig:workflow}
\end{figure*}

Some two-stream based work factorized the video data into content and motion and predict motion separately. For instance, \cite{villegas2017decomposing, denton2017unsupervised} proposed MCnet and Disentangled-representation Net (DRNET) to explicitly separate scene dynamics from the visual appearance. These methods usually assume the visual appearance is static and only predict the dynamics in motion spaces and, as a consequence, they are also limited to one prediction space according to our definition. Lastly, high-level feature spaces can be learned by encoder models, which are usually not semantic explainable anymore \cite{villegas2018hierarchical, denton2018stochastic}. PhyDNet \cite{guen2020disentangling} has been proposed to disentangle PDE dynamics to obtain a latent feature space by utilizing physical knowledge. However, finding a proper high-level feature space is very challenging and the dimensionality reduction of feature space usually comes down to a trade-off between simplicity and quality. 

To date, all the aforementioned approaches attempted to use sophisticated architectures to improve the performance and still limited their models only to one prediction space. In contrast to them, we firstly thoroughly investigate different strategies to construct models that incorporate multi-prediction spaces in this paper.  

%To date, all the existing approaches attempted to use sophisticated architectures to improve the performance and limit their models only to one prediction space. We have found that multi-prediction space-based models have never been fully studied in the literature and in fact, have been largely overlooked so far. In contrast to the previously mentioned methods, we adopt convLSTM \cite{shi2015convolutional} as the base module and thoroughly investigate different strategies to construct models that incorporate multi-prediction spaces.  

\section{Methodology}
\label{sec: model}
In this section, we formulate the mathematical descriptions for the video prediction problem and their architectures. In order to model video data through multi-prediction spaces, there are three important questions to answer: (1) How to obtain the multi-prediction spaces? (2) How to conduct the prediction process on the multi-prediction spaces? (3) How to connect the different prediction spaces? Hereafter, we address these three questions by conducting extensive studies and finally propose a full architecture to deal with this problem.

\subsection{The Prediction Workflow}
Given an input image sequence $\{\mathcal{X}_{1}$, $\mathcal{X}_{2}$, ..., $\mathcal{X}_{T}\}$, our goal is to predict the future frame sequence $\{\mathcal{X}_{T+1}$, $\mathcal{X}_{T+2}$, ..., $\mathcal{X}_{T+N}\}$.
The most common way to video prediction problem is to generate prediction frames in a recursive way, i.e., training a model takes current timestep frame $\mathcal{X}_{t}$ as input to output the next frame $\mathcal{X}_{t+1}$. 
%In training stage, the input is the image $\mathcal{X}_{t}$, and output is $\mathcal{\hat{X}}_{t+1}$, and the loss is given by  $\lVert \mathcal{\hat{X}}_{t+1} - \mathcal{X}_{t+1} \lVert_2$. 
%While in testing stage, the model directly takes the output from previous timestep as input and generates the whole sequence recursively.  
As previously mentioned, prediction methods can be divided into the following three main categories, we illustrate their workflow in Figure \ref{fig:workflow} and present the mathematical formulations:
\vspace{2mm}

\noindent \textbf{Pixel Space Prediction.} The typical prediction method in pixel space can be written as:
\begin{equation}
\mathcal{X}_O = Recurrent\mathcal{M}(\mathcal{X}_I)
\end{equation}
where ${X}_I$ is the input image in time $T$, ${X}_O$ is the output image, and $Recurrent\mathcal{M}$ is usually a RNN based network.
\vspace{2mm}

\noindent \textbf{High-Level Feature Space Prediction.} Encoders and Decoders are employed to translate the input images to the high level feature space $\mathcal{H}$ and back. The corresponding prediction method can be written as:
\begin{align}
\begin{aligned}
\mathcal{H}_I &= Encoder(\mathcal{X}_I) \\
\mathcal{H}_O &= Recurrent\mathcal{M}(\mathcal{H}_I) \\
\mathcal{X}_O &= Decoder(\mathcal{H}_O)
\end{aligned}
\end{align}
where ${\mathcal{H}_I}$ corresponds to the high level feature map obtained by the encoder, ${H}_O$ is the high level feature prediction and the output image ${X}_O$ is obtained through the decoder network. 
\vspace{2mm}

\noindent \textbf{Multi Feature Space Prediction.} In our approach, multiple feature spaces $\mathcal{H}^1, \ \mathcal{H}^2, \ .., \mathcal{H}^k$ for prediction are extracted. Since we believe the incorporation of these feature spaces is able to capture clues of the natural videos in a complementary manner. Therefore, the multi-prediction spaces in our experiment are designed. To be simplified, the equations for a two-prediction spaces scenario can be derived as the followings:

\begin{align}
\begin{aligned}
\mathcal{H}_I &= Encoder(\mathcal{X}_I) \\
\mathcal{H}_O &=  Recurrent\mathcal{M}(\mathcal{H}_I) \\
\mathcal{X}^{'}_O &=  Recurrent\mathcal{M}(\mathcal{X}_I) \\
\mathcal{H}^{'}_O &= Decoder(\mathcal{H}_O) \\
\mathcal{X}_O &= Recurrent\mathcal{M}(Fusion(\mathcal{X}^{'}_O,\mathcal{H}^{'}_O))
\end{aligned}
\end{align}

As illustrated in Figure \ref{fig:workflow}, the two pathways of the prediction process in different feature space are fused together. The precise connection strategy between two prediction spaces is proposed in Section \ref{sec: RC}.

\subsection{The Prediction Spaces}

\begin{table}[b]
	\footnotesize
	\begin{center}
		\setlength{\tabcolsep}{4mm}{
			\begin{tabular}{l|llll}
				\toprule[2pt]
				Model  & MSE & MAE & SSIM & Paras\\
				\midrule[1pt]\midrule[1pt]
				$1^{th}$ layer & 56.3 & 150.7 & 0.816 & 71.2M\\
				$2^{th}$ layer & 34.8 & 98.1 & 0.907 & 70.8M\\
				$3^{th}$ layer & 25.9 & 75.5  & 0.941 & 72.0M\\
				$4^{th}$ layer & 25.7 & 76.1 & 0.942  & 70.7M\\
				All layers & \textbf{18.6} & \textbf{57.4} & \textbf{0.960} & 69.7M\\
				\bottomrule[2pt]
			\end{tabular}
		}
		
	\end{center}
	\caption{The comparison result of using each layer and all layers as prediction space.}
	\label{tab: FeatureSpace}
\end{table}

The most common way to obtain different feature spaces is to operate different scales of convolutions. Usually, the high-level feature space-based methods spend a lot of effort in designing a special prediction space to enhance the performance. However, we discovered through our experiment that even simple feature spaces extracted by typical convolutions show much greater improvements when they are used together.

2D convolutional layers are employed to generate multi-scale features. More detailed information for the layers can be found in the implementation details section. In order to study the characteristic of each feature space, we trained the model,  which adopts ConvLSTM \cite{shi2015convolutional} to predict the future sequences on each single feature spaces and reported their performance and accuracy results here. 

\begin{figure}[t]
	\begin{center}
		\includegraphics[width=1\linewidth]{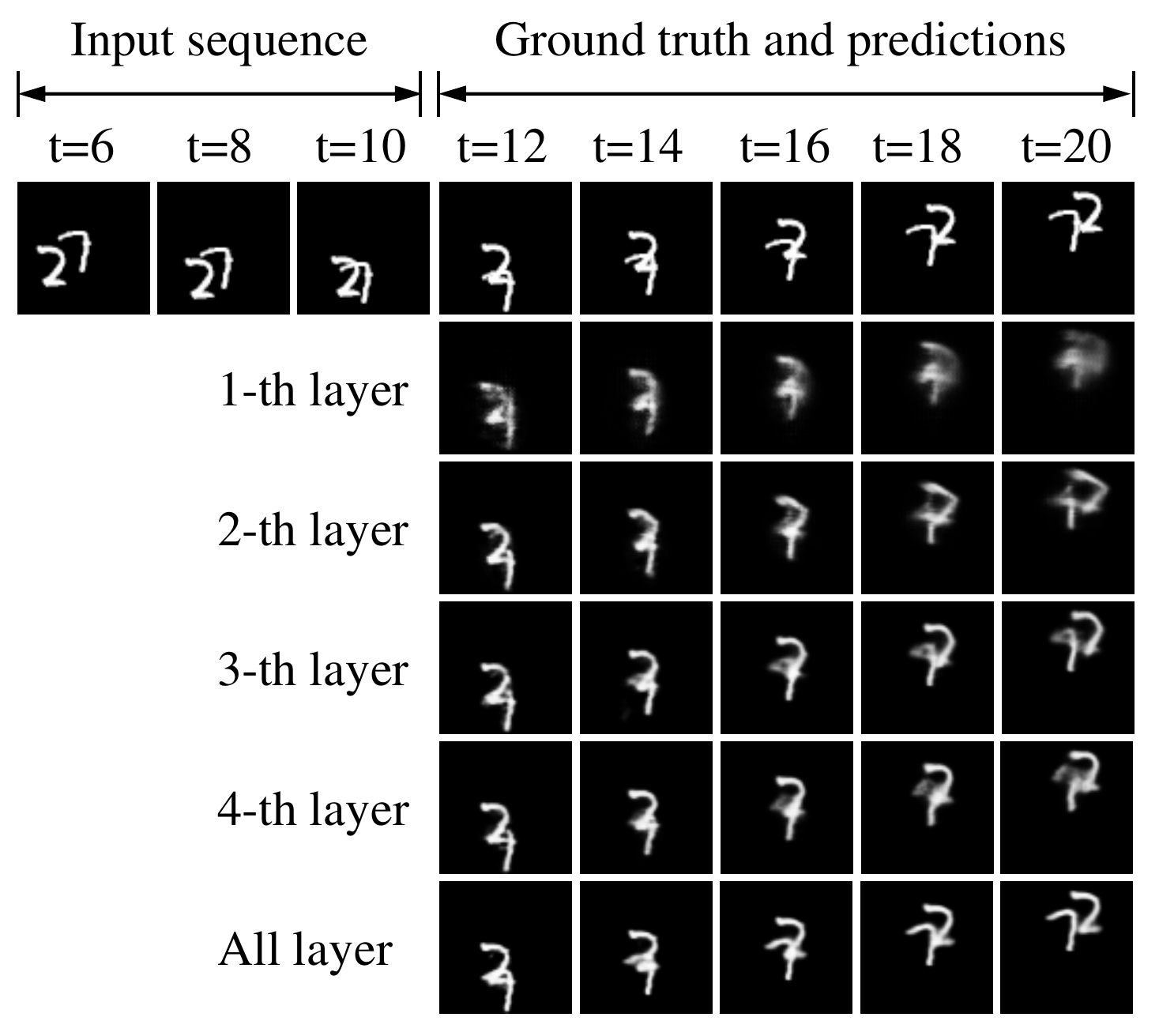}
	\end{center}
	\caption{The comparison results of using each layer as prediction space.}
	\label{fig:FeatureSpace}
\end{figure}

\textbf{Discussion}:
The comparison results presented in Figure \ref{fig:FeatureSpace} and Table \ref{tab: FeatureSpace} show that the deeper feature space is able to preserve the content better without distortions and blurry artefacts. For example, digital numbers four and eight are still readable in the final predicted frame. On the other hand, the numbers are unreadable already from the first frame if the prediction is conducted directly in pixel space. 
Therefore, these observations validate the superiority of the higher-level representation and show that the deeper the feature space is, the higher the accuracy will be.
Most importantly, the model which incorporates the prediction results of all feature spaces achieves the best performance.
These phenomena verify that the comprehensive information of video frames in different prediction spaces can boost the video forecasting tasks.

\subsection{The Recurrent Connection}
\label{sec: RC}
%Common Encoder-Decoder models, like UNet \cite{ronneberger2015u} and UNet++\cite{zhou2018unet++} share a key similarity: skip connections, which combine the deep, semantic, coarse-grained feature map of the decoder sub-network with shallow, low-level, fine-grained feature map of the encoder sub-network. Skip connections have effectively been able to provide fine-grained features, which are much more important for target image generation. Traditional skip connections are used for segmentation tasks, in which the source image and target image have no spatial-temporal differences. 

The key issue to perform our idea is how to connect different prediction processes to fuse the generations from multi-feature space. Therefore, a novel Recurrent Connection strategy is designed in this paper to address this problem. Assuming the input frame at time $t$ and layer $l$ is denoted as $\mathcal{H}^{l}_{t}$, and the predicted feature on the next timestep and layer $l$ is denoted as $\mathcal{H}^{l}_{t+1}$. As a consequence, there is a spatial-temporal gap between these two features so that the core technologies are proposed to model the transition from $\mathcal{H}^{l}_{t}$ to $\mathcal{H}^{l}_{t+1}$. As shown in Figure \ref{fig:workflow}, the methods of pixel space prediction models this transition by a recurrent cell such as ConvLSTM. In our model, a recurrent connection strategy will connect the prediction features every two layers and construct a new way to model the transition from $\mathcal{H}^{l}_{t}$ to $\mathcal{H}^{l}_{t+1}$.

%However, for video prediction tasks, the features of the encoder network are from the image at time $t$, whereas the features of the decoder network belong to time $t+1$. As a consequence, there is a spatial-temporal gap between these two features. In more detail, the features from the encoder network need to be transformed before the fusion with the features of the decoder network happens. To address this issue, we propose recurrent skip connections (RSC), which use a LSTM block to accomplish the transformation of the encoder features. The structure of our RSC is shown in Figure \ref{fig:U}. Three popular LSTM-based models for sequences prediction are employed in our experiments, namely ConvLSTM \cite{shi2015convolutional}, ConvGRU \cite{shi2017deep} and PredRNN \cite{wang2017predrnn}. The comparison of the different types of recurrent connections is reported in the experiment section. 

The detailed design of the propagation of sequential data in our Recurrent Connection can be described as:

%\begin{align}
%\begin{aligned}
%     & \hat{\mathcal{H}}^{l}_{t} &&= ConvLSTM(\mathcal{H}^{l}_{t}) \\
%     & \mathcal{H}^{l+1}_{t} &&= DownSample(\hat{\mathcal{H}}^{l}_{t}) \\
%     & \hat{\mathcal{H}}^{l}_{t+1} &&= ConvLSTM(\hat{\mathcal{H}}^{l}_{t}) \\
%     & \hat{\mathcal{H}}^{l+1}_{t+1} &&= ConvLSTM(\hat{\mathcal{H}}^{l+1}_{t}) \\
%     & \mathcal{H}^{l+1}_{t+1} &&= ConvLSTM (\hat{\mathcal{H}}^{l+1}_{t}) \\
%     & X_{t+1} &&= ConvLSTM(Fusion(\mathcal{H}^{l}_{t+1},UpSample(\mathcal{H}^{l+1}_{t+1})))
%\end{aligned}
%\end{align}

\begin{align}
\begin{aligned}
& \hat{\mathcal{H}}^{l}_{t} &&= ConvLSTM(\mathcal{H}^{l}_{t}) \\
& \mathcal{H}^{l+1}_{t} &&= DownSample(\hat{\mathcal{H}}^{l}_{t}) \\
& \hat{\mathcal{H}}^{l}_{t+1} &&= ConvLSTM(\hat{\mathcal{H}}^{l}_{t}) \\
& \mathcal{H}_{t+1}^l &&= ConvLSTM(Fusion(\hat{\mathcal{H}}^{l}_{t+1},UpSample(\mathcal{H}^{l+1}_{t+1})))
\end{aligned}
\end{align}

In the equation, $\mathcal{H}^{l}_t$ and $\mathcal{H}^{l}_{t+1}$ corresponds to the propagation in the $l^{th}$ layer . $\mathcal{H}^{l+1}_t$ and $\mathcal{H}^{l+1}_{t+1}$ represents the propagation in the $(l+1)^{th}$ layer. The feature $\mathcal{H}_{t+1}^{l+1}$ is obtained from Recurrent Connection between the $(l+1)^{th}$ layer and the $(l+2)^{th}$ layer. The last layer of the architecture will perform the transition of frames by typical recurrent cells. Two LSTM cell are employed before and after the DownSample and UpSample process to strengthen the ability to capture the spatio-temporal clues. Three types of RNN block for sequences prediction are employed in our experiments, namely ConvLSTM \cite{shi2015convolutional}, ConvGRU \cite{shi2017deep} and PredRNN \cite{wang2017predrnn}. After consideration of efficiency and performace, we use ConvLSTM in our final model. 

\noindent \textbf{Fusion Strategy.} For every input frame $\mathcal{X}_{t}$, there will be one frame $\mathcal{X}^{'}_{t+1}$ generated from the current prediction space and another frame $\mathcal{X}^{''}_{t+1}$ obtained from the upsampling of the lower-level prediction space. It is quite important to investigate how the low-level prediction feature map should be fused with the high-level prediction feature map. Various fusion functions are proposed, we chose the most widely used four types of fusion functions \cite{feichtenhofer2016convolutional} , namely  Sum Fusion, Concatenation Fusion, Max Fusion and Attention Fusion. Comparison results are reported in Section \ref{exp}.

%Assume the input feature maps are $x'_{T+1} \in \mathbb{R}^{H1 \times W1 \times C1}$ and $H_{T+1} \in \mathbb{R}^{H2 \times W2 \times C2} $, a fusion function $f$ will generated a new feature map $x_{T+1} \in \mathbb{R}^{H3 \times W3 \times C3}$. For training purpose, only the differential-able fusion function is considered. A various fusion functions are proposed before, we chose the most widely used three type of fusion function, Sum Fusion, Concatenation Fusion and Attention Fusion.

\section{Experiments}
\label{exp}

In this section, we will first introduce the datasets used and implementation details of our model. Then we will verify the performance of our model by comparing it with the five competitive models including ConvLSTM \cite{shi2015convolutional}, PredRNN \cite{wang2017predrnn}, MIM \cite{wang2019memory}, CrevNet \cite{yu2019efficient} and PhyDNet \cite{guen2020disentangling}. %Furthermore, inspired by the previous revolution of RNN blocks (ConvLSTM \cite{shi2015convolutional},ST-LSTM \cite{wang2017predrnn}), we explore different variations of RNN blocks. Finally, we show that our recurrent connection is generic and can be used to improve the performance of other video prediction architectures. 

\noindent \textbf{Implementation details.} Our model is trained by using $\ell_{2}$ loss and ADAM \cite{kingma2014adam} optimizer with an initial learning rate of $0.0005$. The kernel size of all convolutional layers and ConvLSTMs is set to $3 \times 3$. 
The RNN path is composed of a 1-layer ConvLSTM with the same number of filters as its input channels. 
All the experiments are implemented in PyTorch \cite{paszke2019pytorch} and conducted on NVIDIA V100 Tensor Core GPUs.

\noindent \textbf{Evaluation metrics.} We use three common evaluation metrics used in previous video prediction works: Mean Squared Error (MSE), Mean Absolute Error (MAE) and Structural Similarity (SSIM \cite{wang2004image}). These metrics are averaged for each frame of the output sequence. Lower values of MSE, MAE and higher values for SSIM indicate better performances.

\subsection{Datasets}
\noindent \textbf{Moving MNIST} \cite{srivastava2015unsupervised} is a standard benchmark in video prediction containing 2 random moving digits bouncing inside a 64 $\times$ 64 grid. Each sequence contains 20 frames, 10 for the input sequences and 10 for the prediction outcome. We generate Moving MNIST sequences following the method in \cite{wang2017predrnn} for training and use the fixed test set of 10,000 sequences provided by \cite{srivastava2015unsupervised} for evaluation.

\noindent \textbf{KTH} \cite{schuldt2004recognizing} includes 6 kinds of human actions (walking, jogging, running, boxing, hand waving and hand clapping) performed in different scenarios. The video frames used in our experiment are subsampled from the original videos at 1Hz, and resized to be 128 $\times$ 128. All video frames are divided into a training set (subjects 1-16) and a test set (subjects 17-25). Following the common setup for this benchmark, we generate 10 frames from the last 10 observations. In our experiments, we use 172,372 sequences for training and 99,375 sequences for testing.

\noindent \textbf{Human 3.6M} \cite{ionescu2013human3} contains 17 kinds of human actions, including 3.6 million poses and corresponding images. The original images in Human3.6M dataset are 1000 $\times$ 1000  $\times$ 3. All images are centered and cropped to 800 $\times$ 800 $\times$ 3, and resized to 128 $\times$ 128 $\times$ 3 in our experiments. 
According to \cite{wang2019memory}, subjects S1, S5, S6, S7, S8 are used for training and S9, S11 used for testing. We evaluate 4 future frames given the previous 4 input frames.

\noindent \textbf{Radar Echo dataset} is generated from historical radar maps from local weather with an interval of 6 minutes. Each frame is a 700 $\times$ 900  $\times$ 1 grid for an image, covering 700 $\times$ 900 square kilometers. To effectively train all the models for comparison, the images are firstly converted to 900 $\times$ 900 $\times$ 1 with zero paddings and then resized to  128 $\times$ 128 $\times$ 1 for training and testing. We predict 10 radar maps at a time interval of 6 minutes, covering the next hour. In our experiments, we use 9,600 radar sequences for training and 2,400 for testing.

\begin{table}[t]
	\footnotesize
	\begin{center}
		\setlength{\tabcolsep}{1.5mm}{
			\begin{tabular}{l|lll|lll}
				\toprule[2pt]
				&   \multicolumn{3}{c|}{MNIST-2}  & \multicolumn{3}{c}{MNIST-3} \\
				\midrule[1pt]
				Model    & MSE & MAE & SSIM & MSE & MAE & SSIM \\
				\midrule[1pt]\midrule[1pt]
				ConvLSTM \cite{shi2015convolutional} & 103.3 & 182.9 & 0.707  & 127.3 & - & 0.695    \\
				PredRNN \cite{wang2017predrnn} & 56.8 & 126.1 & 0.867  & 83.1  & - & 0.822    \\
				MIM \cite{wang2019memory} & 44.2 & 101.1 & 0.910  & 49.8* & 138.2*  & 0.879*  \\
				CrevNet \cite{yu2019crevnet} & 22.3 & - & 0.949  & 40.6 & - &    0.916  \\
				PhyDNet \cite{guen2020disentangling} & 24.4 & 70.3 & 0.947  & 50.8* & 136.4*  & 0.888*     \\
				Ours & \textbf{17.2} & \textbf{47.7} & \textbf{0.967}    & \textbf{29.8} &\textbf{80.1}  &\textbf{0.944} \\
				\bottomrule[2pt]
			\end{tabular}
		}
	\end{center}
	
	\caption{The comparison result of our model with other State-of-the-Art models on Moving MNIST-2 and Moving MNIST-3 datasets. * corresponds to results obtained by running the online code from the authors.
	}
	\label{MNIST}
\end{table}

\subsection{Comparison with the State-of-the-Art work}

We report here the evaluation of our proposed strategies against leading research works in video prediction. In our final model, four layers of feature spaces and ConvLSTM are used. The top performance fusion function is selected as well.

%We evaluated our strategies with respect the leading research works in video prediction, including the most representative five baselines: ConvLSTM \cite{shi2015convolutional}, MIM \cite{wang2019memory}, PredRNN \cite{wang2017predrnn} and PhyDNet\cite{guen2020disentangling}. In our final model, four layers of feature spaces and ConvLSTM are used. The top performance fusion function is selected as well. 
%The six common metrics on the Moving MNIST dataset is evaluated first and the typical three metrics are evaluated on the other three benchmarks. 

\begin{figure}[b]
	\begin{center}
		\includegraphics[width=1\linewidth]{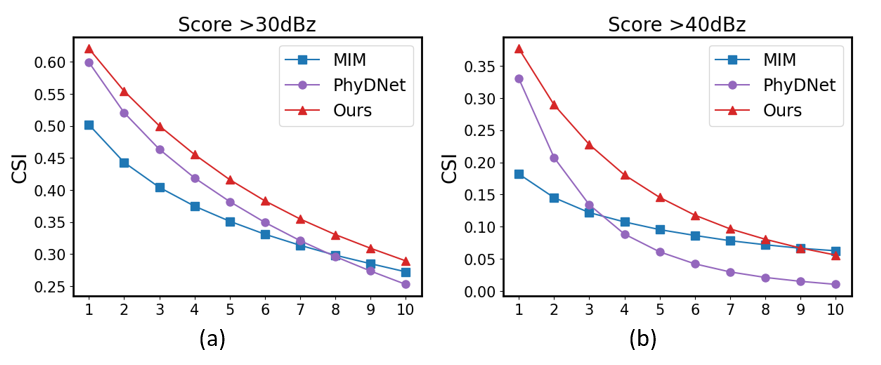}
	\end{center}
	\caption{Frame-wise comparisons of the next 10 generated radar
		maps. Higher CSI curves indicate better forecasting results. }
	\label{fig:CSI}
\end{figure}

\textbf{Quantitative} As shown in Table \ref{MNIST} and \ref{SOTA}, our proposed model achieves marginal improvements in comparison with all the five baselines on all the datasets and terms, therefore this evidence validates the effectiveness of the idea to use multi-prediction spaces. 
From the results on the Moving MNIST with 3 handwritten digits, almost half of the value of MSE and MAE are drastically decreased when compared to the competitive PhyDNet model. It is worth noting that this big gap had not been introduced by the various previously proposed approaches. 
Moreover, our model outperforms the State-of-the-Art method significantly on the other three real-world video benchmarks. On the challenging Radar Echo Dataset, our model yields the best results (76.7 MSE, 319.3 MAE and 0.942 SSIM) to date. 
Moreover, the large and persistent performance boost is reported in the term of SSIM, indicating that our model also has superior ability in preserving the image quality during the prediction task. 
In addition, we reported the CSI scores \cite{xingjian2015convolutional} specific for weather forecasting task in Figure \ref{fig:CSI}.
Our model consistently outperforms the compared methods in the future 10 frames (1 hour),
which indicates the superior performance of our model on the precipitation nowcasting task.

\begin{figure}[t]
	\begin{center}
		\includegraphics[width=1\linewidth]{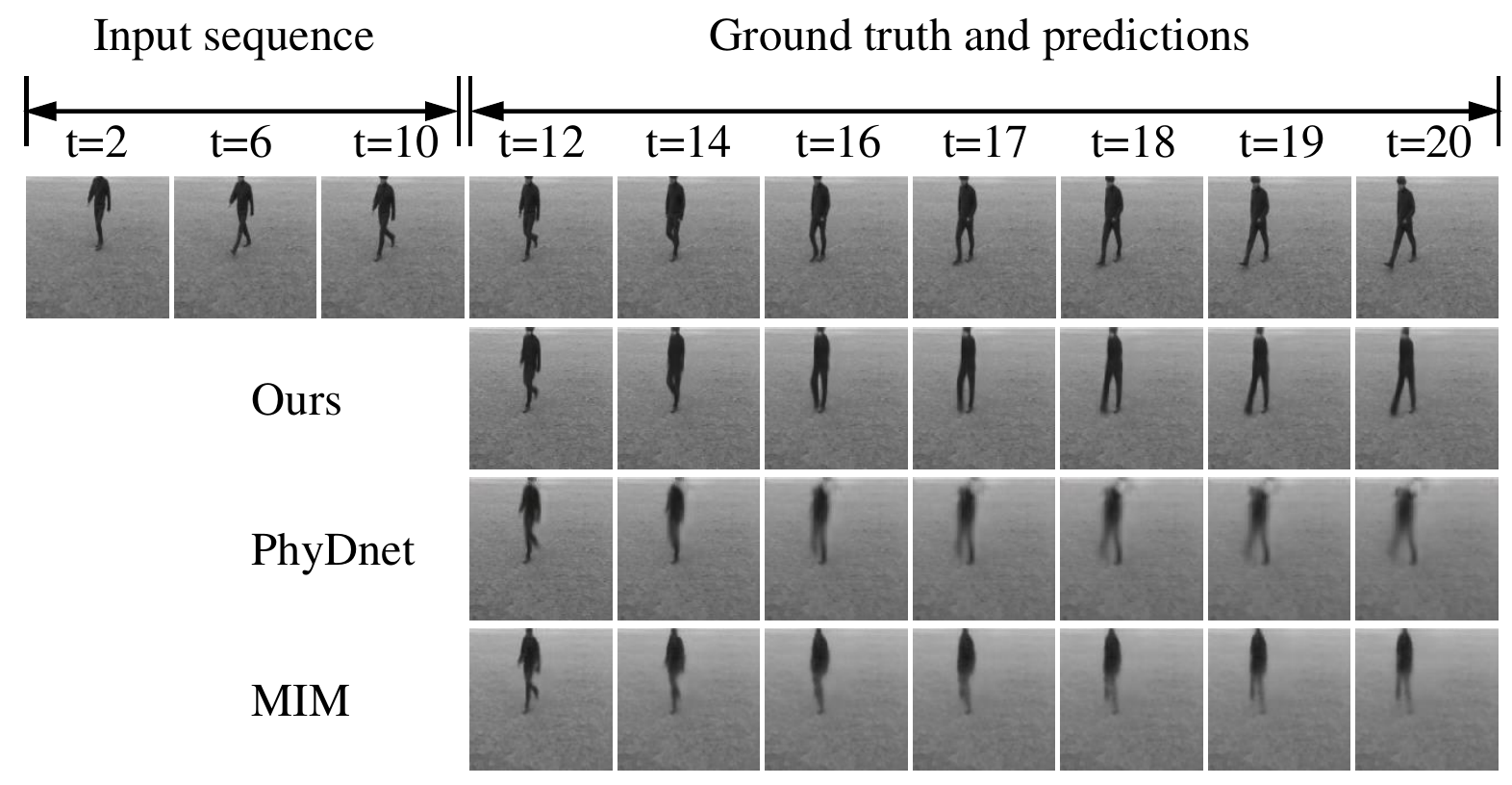}
	\end{center}
	\caption{Prediction examples on the KTH dataset of our model and the State-of-the-Art works. The MIM \cite{wang2019memory} model predicts the future frames on pixel space and the PhyDNet \cite{guen2020disentangling} model conducts the prediction on high-level feature space.}
	\label{fig:KTHsota}
\end{figure}

\begin{figure}[t]
	\begin{center}
		\includegraphics[width=1\linewidth]{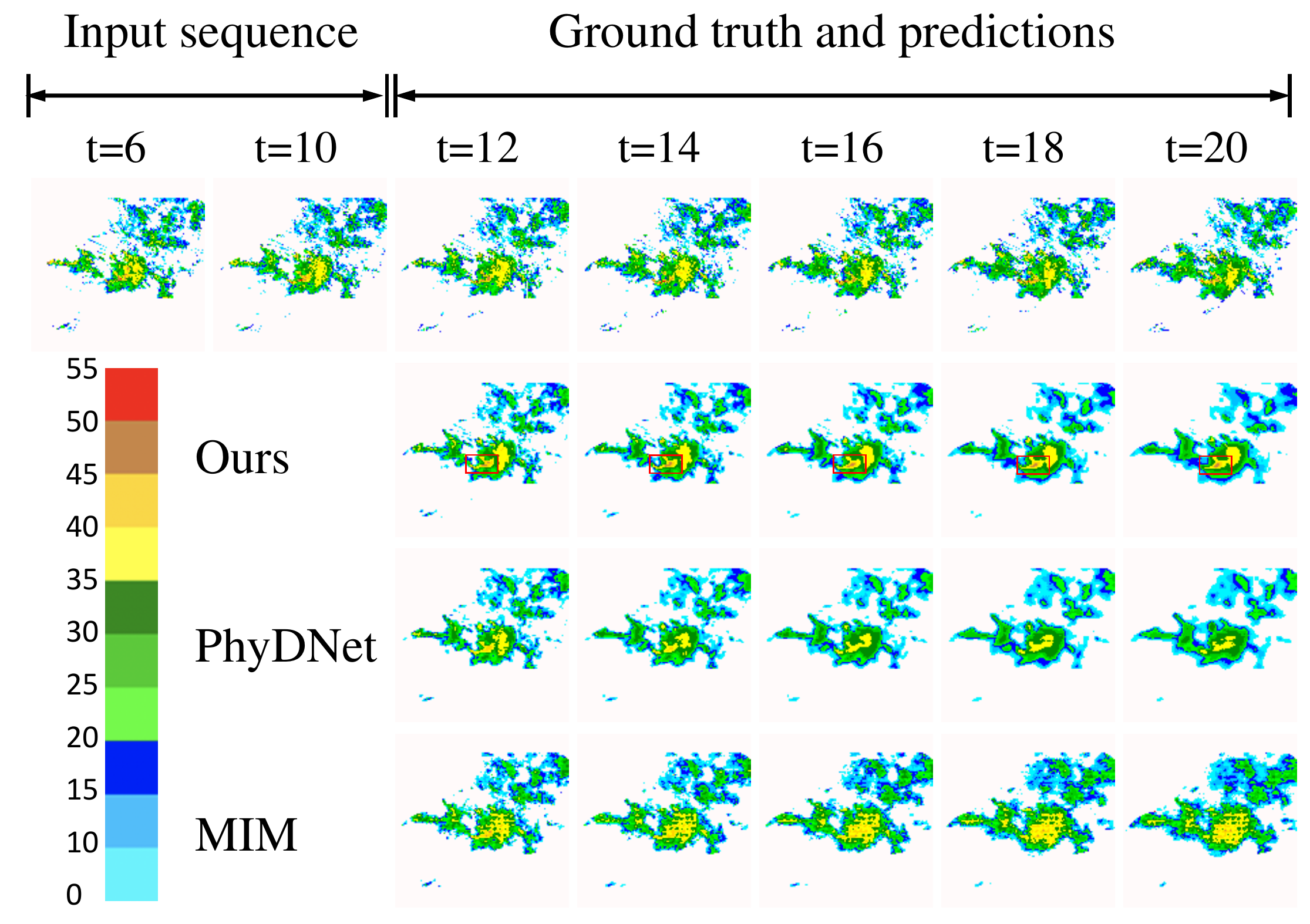}
	\end{center}
	\caption{ Prediction examples on the Radar Echo Dataset of our model and the State-of-the-Art works. }
	\label{fig:weather}
\end{figure}

\begin{figure}[t]
	\begin{center}
		\includegraphics[width=0.95\linewidth]{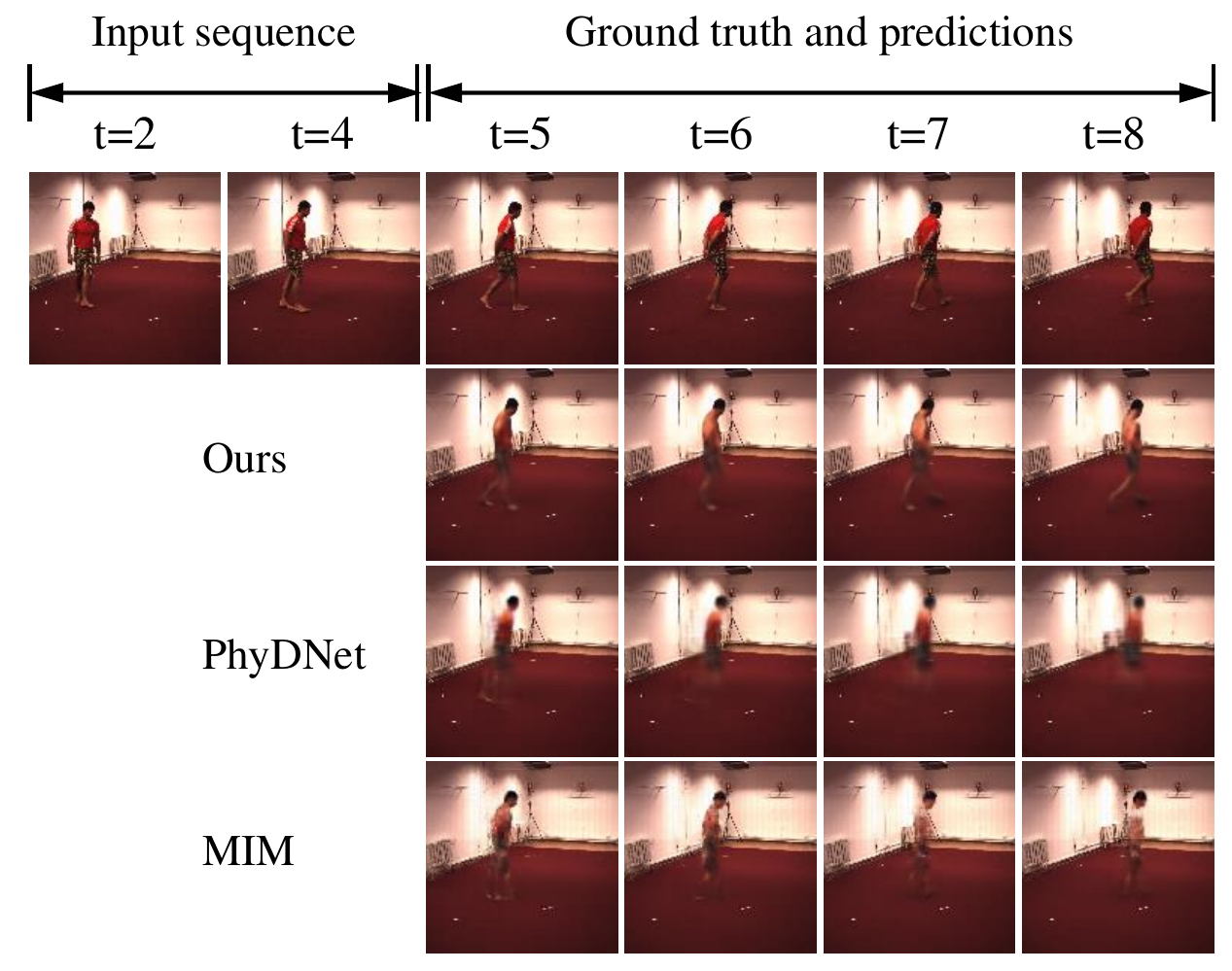}
	\end{center}
	\caption{ Prediction examples on the human action videos of our model and the State-of-the-Art works. }
	\label{fig:HumanAction}
\end{figure}

\textbf{Qualitative} The qualitative comparison results of our model and the State-of-the-Art method is illustrated in Figure \ref{fig:KTHsota}, \ref{fig:weather}, \ref{fig:HumanAction}. 

We can observe from Figure \ref{fig:KTHsota} that the proposed model can clearly predict the pose of human and their position,
while blurry and incorrect results derive from the other two models.

For the Radar Echo Dataset, different colors represent different radar dBZ values, and high dBZ value is more important for weather nowcasting. As shown in Figure \ref{fig:weather}, out model can successfully preserve the high dBZ values (the area in the red blocks) in the prediction frames, our prediction results are much sharper than other models, which is rather useful to get more accurate weather forecasting. 

Furthermore, the prediction results of Human action videos presented in Figure \ref{fig:HumanAction}, compared with the State-of-the-Art method, surprisingly show that our multi-prediction spaces method is more advantageous -- the human legs are still distinct and accurate in the predicted future.

All the visualization results validate the remarkable advantages of the idea to leverage both the high-level semantic information and low-level pixel information by using the recurrent connection mechanism. More visualization comparisons on all the datasets are provided in our Demo and Supplementary materials.

\begin{table*}[!]
	\footnotesize
	\begin{center}
		\setlength{\tabcolsep}{4.5mm}{
			\begin{tabular}{l|ccc|ccc|ccc}
				\toprule[2pt]
				&\multicolumn{3}{c|}{KTH}  & \multicolumn{3}{c|}{Radar Echo Dataset} & \multicolumn{3}{c}{Human 3.6M}\\
				\midrule[1pt]
				Model  & MSE & MAE & SSIM  & MSE & MAE & SSIM & MSE & MAE & SSIM \\
				\midrule[1pt]\midrule[1pt]
				PredRNN \cite{wang2017predrnn} & 180.3 & 1985.1 & 0.831   & 92.5  & 365.3  & 0.737 & 481.4  & 1895.2 &  0.781 \\
				MIM \cite{wang2019memory} & 122.2 & 1347.3 & 0.840   & 92.9  & 402.6  & 0.725 &  429.9 & 1782.8 & 0.790 \\
				PhyDNet \cite{guen2020disentangling} & 91.2 & 1061.1 & 0.902 & 41.5 & 280.1 & 0.859 & 264.6 & 1129.2 & 0.929 \\
				Ours & \textbf{75.9} & \textbf{834.0} & \textbf{0.924}   &  \textbf{39.7}&  \textbf{241.7} &  \textbf{0.872} & \textbf{230.1}  & \textbf{947.9}  &  \textbf{0.942}   \\
				\bottomrule[2pt]
		\end{tabular}}
	\end{center}
	\caption{The comparison result of our model with other State-of-the-Art models on the KTH, Human 3.6M and Radar Echo Datasets. Our model is the most high-performing method over all datasets.}
	\label{SOTA}
\end{table*}

\section{Ablation Study}
Extensive ablation studies are carried out in this section to thoroughly investigate the best way to design the whole architecture, including the analysis of each component of our model. In the following, we address and deeply explore the three questions discussed in Section \ref{sec: model}.

\subsection{The Number of Feature Layers}

\begin{figure}[!]
	\begin{center}
		\includegraphics[width=1\linewidth]{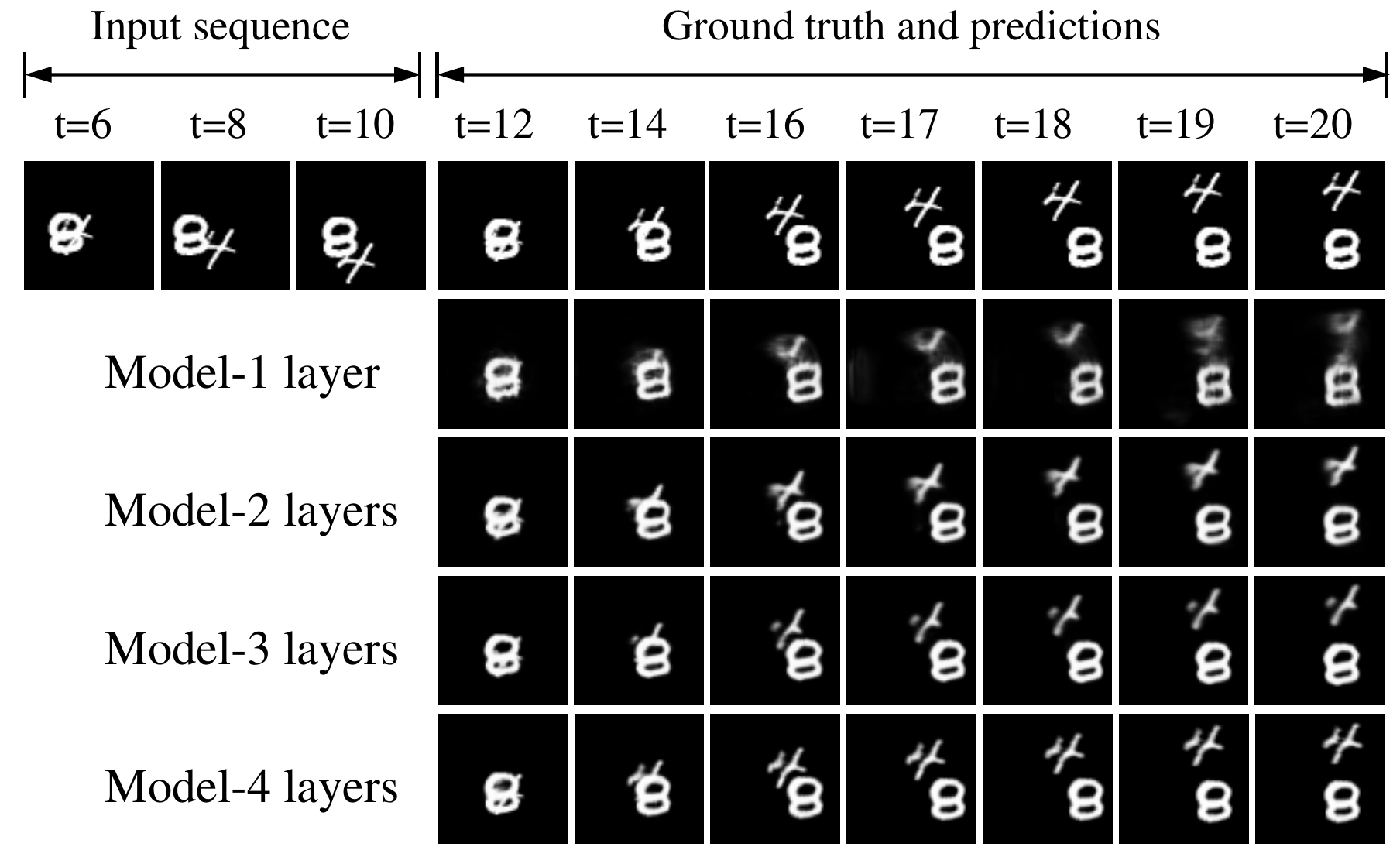}
	\end{center}
	\caption{The comparison results of using different number of prediction spaces.}
	\label{fig:Deep0fModel}
\end{figure}

As shown in Table \ref{Ablation} (a), We evaluated four different baselines on the Moving MNIST dataset to study the relationship between the number of prediction spaces and the accuracy. 
The $model$-$N$ baseline denotes the model with $N$ layers.
It can be clearly observed that the utilization of more prediction spaces will lead to more favourable performance gains. While $model$-$1$ just perform prediction on the single Pixel Space, the involvement of the second layer of feature space brings significant improvements, with a drop by $31.9$ and $73.0$ in terms of MSE and MAE respectively. From the results of $model$-$3$ and $model$-$4$, we can see the errors decrease slower however the improvements in performance will be harder at that level as well. When the prediction spaces are increased from one to four, we see a remarkable drop on MSE by 42.6 and on MAE by 104.5. Concurrently to the continuous enhancements on MSE and MAE, the quality of the predicted frames is also increasing significantly according to the SSIM metric. This also can be seen from the illustration in Figure \ref{fig:Deep0fModel}. Based on all these observations, we can confidently conclude from the experiments results that increasing the number of prediction spaces will introduce further performance gains to the video forecasting task.

\subsection{The Fusion Strategy}
The other question we want to investigate is how the different fusion functions will contribute to the final results. Here we set the same number of prediction spaces and compare the four most commonly used fusion strategies \cite{feichtenhofer2016convolutional}, namely sum fusion, concatenate fusion, max fusion and attention fusion.

From the report in Table \ref{Ablation} (b), the Max fusion function achieved the top performance among the four on the term of MSE. It surpasses the concatenate fusion function by a gap of 1.9. However, the concatenate fusion function achieves the best performance in terms of MAE. A slightly lower error of 47.7 is obtained by concatenate fusion. On the last column of Table \ref{Ablation} (b), we can see that the SSIM values of the three fusion functions is almost the same. Therefore this evaluation suggests that all the different types of fusion strategies can achieve very similar results and more complex fusion function as attention does not lead to significantly better results. 
\subsection{The Recurrent Connection Module}

Additionally, we carefully studied how the design of recurrent connections will affect the performance of the final model. To accomplish this, we selected the competitive recurrent models, i.e. ConvLSTM \cite{shi2015convolutional}, ConvGRU \cite{ballas2015delving}  and ST-LSTM \cite{wang2017predrnn} for spatial-temporal data prediction.

The following observations can be found in Table \ref{Ablation} (c). Firstly, the ConvLSTM module outperforms the other two modules in a small gap, which is 1.0 MSE less than ST-LSTM and 4.4 MSE less than ConvGRU. %Interestingly, ST-LSTM and ConvGRU are usually reported to be more effective in a single prediction space. 
The same conclusion can be deduced from the comparison of MAE in the second column. Secondly, the differences of the three baselines are quite small when they are compared with the sharper difference of 31.9 MSE by increasing new prediction spaces, therefore these results clearly underline the dominant importance of the proposed idea Single-To-Multiple. 

\begin{table}
	\footnotesize
	\begin{center}
		\setlength{\tabcolsep}{3mm}{
			\begin{tabular}{c|l|ccc}
				\toprule[1.5pt]
				&Model    & MSE & MAE & SSIM \\
				\midrule[1pt]\midrule[1pt]
				\multirow{4}{*}{(a)} &Model-1  & 59.8 & 152.2 & 0.816     \\
				&Model-2  & 27.9 & 79.2 & 0.935  \\
				&Model-3  & 19.2 & 58.7 & 0.958 \\
				&Model-4  & \textbf{17.2} & \textbf{47.7} & \textbf{0.967}    \\
				\midrule[1.5pt]
				\multirow{4}{*}{(b)}&Sum & 15.4 & 48.6 & \textbf{0.967}   \\
				&Attention& 16.1 & 49.6 & 0.966 \\
				&Concatenate & 17.2 & \textbf{47.7} & \textbf{0.967}  \\
				&Max & \textbf{15.3} & 48.3 & \textbf{0.967}   \\
				\midrule[1.5pt]
				\multirow{3}{*}{(c)}&Ours with ConvGRU & 21.6 & 63.7 & 0.953     \\
				&Ours with ST-LSTM & 18.2 &  48.8 & 0.966  \\
				&Ours with ConvLSTM & \textbf{17.2} &  \textbf{47.7} & \textbf{0.967} \\
				\bottomrule[1.5pt]
			\end{tabular}
		}
	\end{center}
	\caption{Quantitative ablation results of our model. Part (a) shows the influence of the number of prediction space. Part (b) shows the sensitivity of of our model to the different fusion way. Part (c) shows the influence of different RNN cells in our recurrent connection module. }
	\label{Ablation}
\end{table}

\subsection{Generalization of our idea}
%The above experiments have sufficiently demonstrated the effectiveness of the incorporation of different scales of convolution operated feature spaces. 
Finally, a natural question arises, that is if this idea could also be applicable to other types of high-level feature space? Motivated by this, we evaluated our idea by employing another base module called PhyDNet \cite{guen2020disentangling}, 
which is a typical encoder-predictor-decoder framework.
The feature maps from the layers of the encoder are propagated to the layer of the decoder directly, in a similar way as to the U-Net model. However, there is no prediction process during this link on a different scale of feature maps. Therefore, we integrated and implemented our idea by revising the propagation so that multi-prediction processes are introduced in the model. Comparison results in terms of MSE, MAE and SSIM are presented in Table \ref{Table:Generalization} show that considerable gains can be achieved to use this idea on the feature spaces of PhyDNet as well. Therefore, we can conclude that the evaluation study indicates that the key idea of this paper generalizes well for diverse types of feature spaces to achieve significant improvements.    
\begin{table}[t]
	\footnotesize
	\begin{center}{
			\setlength{\tabcolsep}{4.5mm}{
				\begin{tabular}{l|lll}
					\toprule[1.5pt]
					Model    & MSE & MAE & SSIM \\
					\midrule[1pt]\midrule[1pt]
					PhyDNet \cite{guen2020disentangling}  & 33.2* & 90.6* & 0.925*     \\
					PhyDNet+Ours  & 26.5 & 75.5 & 0.942    \\
					\bottomrule[1.5pt]
			\end{tabular}}
		}
	\end{center}
	\caption{Generalization to other modalities. *results obtained by running online code from the authors, which are different from the scores reported in its original paper[\cite{guen2020disentangling}]. We tried to contact the author, but there was no response until submitted.
	}
	\label{Table:Generalization}
\end{table}

\section{Conclusion}
In this paper, to the best of our knowledge, we are the first to systematically investigate the largely neglected yet significant strategy to introduce multi-prediction spaces for the task of video forecasting. This is unlike the more common approach of just using a single prediction space. However, the results obtained through our extensive experiments on four datasets, indicate that our proposed idea is capable of drastically boosting the performance over the State-of-the-Art methods. We conducted a series of ablation studies on various ways to factorize the feature spaces, fuse the feature maps from different prediction spaces, and build up Recurrent Connection. Furthermore, evaluations on other prediction spaces demonstrate the generalization of our idea. Finally, we also found that the prediction results from pixel space and high-level feature spaces are complementary to each other and can be fused together to futher improve the performance. This study is a first step towards enhancing our understanding of this topic, and we believe that the various in-depth component analysis will provide a lot of important insights that will hopefully encourage the future research work in the video forecasting domain.
\newpage

{\small
	\bibliographystyle{ieee_fullname}
	\bibliography{egbib}

\begin{thebibliography}{10}\itemsep=-1pt

\bibitem{ballas2015delving}
Nicolas Ballas, Li Yao, Chris Pal, and Aaron Courville.
\newblock Delving deeper into convolutional networks for learning video
  representations.
\newblock {\em arXiv preprint arXiv:1511.06432}, 2015.

\bibitem{chiu2020segmenting}
Hsu-kuang Chiu, Ehsan Adeli, and Juan~Carlos Niebles.
\newblock Segmenting the future.
\newblock {\em IEEE Robotics and Automation Letters}, 5(3):4202--4209, 2020.

\bibitem{denton2017unsupervised}
Emily Denton and Vighnesh Birodkar.
\newblock Unsupervised learning of disentangled representations from video.
\newblock {\em arXiv preprint arXiv:1705.10915}, 2017.

\bibitem{denton2018stochastic}
Emily Denton and Rob Fergus.
\newblock Stochastic video generation with a learned prior.
\newblock In {\em International Conference on Machine Learning}, pages
  1174--1183. PMLR, 2018.

\bibitem{feichtenhofer2016convolutional}
Christoph Feichtenhofer, Axel Pinz, and Andrew Zisserman.
\newblock Convolutional two-stream network fusion for video action recognition.
\newblock In {\em Proceedings of the IEEE conference on computer vision and
  pattern recognition}, pages 1933--1941, 2016.

\bibitem{finn2016unsupervised}
Chelsea Finn, Ian Goodfellow, and Sergey Levine.
\newblock Unsupervised learning for physical interaction through video
  prediction.
\newblock In {\em Advances in neural information processing systems}, pages
  64--72, 2016.

\bibitem{guen2020disentangling}
Vincent~Le Guen and Nicolas Thome.
\newblock Disentangling physical dynamics from unknown factors for unsupervised
  video prediction.
\newblock In {\em Proceedings of the IEEE/CVF Conference on Computer Vision and
  Pattern Recognition}, pages 11474--11484, 2020.

\bibitem{hou2019video}
Ruibing Hou, Hong Chang, Bingpeng Ma, and Xilin Chen.
\newblock Video prediction with bidirectional constraint network.
\newblock In {\em 2019 14th IEEE International Conference on Automatic Face \&
  Gesture Recognition (FG 2019)}, pages 1--8. IEEE, 2019.

\bibitem{hu2019novel}
Zhihang Hu and Jason Wang.
\newblock A novel adversarial inference framework for video prediction with
  action control.
\newblock In {\em Proceedings of the IEEE International Conference on Computer
  Vision Workshops}, pages 0--0, 2019.

\bibitem{ionescu2013human3}
Catalin Ionescu, Dragos Papava, Vlad Olaru, and Cristian Sminchisescu.
\newblock Human3. 6m: Large scale datasets and predictive methods for 3d human
  sensing in natural environments.
\newblock {\em IEEE transactions on pattern analysis and machine intelligence},
  36(7):1325--1339, 2013.

\bibitem{jin2017video}
Xiaojie Jin, Xin Li, Huaxin Xiao, Xiaohui Shen, Zhe Lin, Jimei Yang, Yunpeng
  Chen, Jian Dong, Luoqi Liu, Zequn Jie, et~al.
\newblock Video scene parsing with predictive feature learning.
\newblock In {\em Proceedings of the IEEE International Conference on Computer
  Vision}, pages 5580--5588, 2017.

\bibitem{jin2017predicting}
Xiaojie Jin, Huaxin Xiao, Xiaohui Shen, Jimei Yang, Zhe Lin, Yunpeng Chen,
  Zequn Jie, Jiashi Feng, and Shuicheng Yan.
\newblock Predicting scene parsing and motion dynamics in the future.
\newblock In {\em Advances in Neural Information Processing Systems}, pages
  6915--6924, 2017.

\bibitem{kalchbrenner2017video}
Nal Kalchbrenner, A{\"a}ron Oord, Karen Simonyan, Ivo Danihelka, Oriol Vinyals,
  Alex Graves, and Koray Kavukcuoglu.
\newblock Video pixel networks.
\newblock In {\em International Conference on Machine Learning}, pages
  1771--1779. PMLR, 2017.

\bibitem{kim2019unsupervised}
Yunji Kim, Seonghyeon Nam, In Cho, and Seon~Joo Kim.
\newblock Unsupervised keypoint learning for guiding class-conditional video
  prediction.
\newblock In {\em Advances in Neural Information Processing Systems}, pages
  3814--3824, 2019.

\bibitem{kingma2014adam}
Diederik~P Kingma and Jimmy Ba.
\newblock Adam: A method for stochastic optimization.
\newblock {\em arXiv preprint arXiv:1412.6980}, 2014.

\bibitem{kwon2019predicting}
Yong-Hoon Kwon and Min-Gyu Park.
\newblock Predicting future frames using retrospective cycle gan.
\newblock In {\em Proceedings of the IEEE Conference on Computer Vision and
  Pattern Recognition}, pages 1811--1820, 2019.

\bibitem{lotter2016deep}
William Lotter, Gabriel Kreiman, and David Cox.
\newblock Deep predictive coding networks for video prediction and unsupervised
  learning.
\newblock {\em arXiv preprint arXiv:1605.08104}, 2016.

\bibitem{luc2017predicting}
Pauline Luc, Natalia Neverova, Camille Couprie, Jakob Verbeek, and Yann LeCun.
\newblock Predicting deeper into the future of semantic segmentation.
\newblock In {\em Proceedings of the IEEE International Conference on Computer
  Vision}, pages 648--657, 2017.

\bibitem{mathieu2015deep}
Michael Mathieu, Camille Couprie, and Yann LeCun.
\newblock Deep multi-scale video prediction beyond mean square error.
\newblock {\em arXiv preprint arXiv:1511.05440}, 2015.

\bibitem{minderer2019unsupervised}
Matthias Minderer, Chen Sun, Ruben Villegas, Forrester Cole, Kevin~P Murphy,
  and Honglak Lee.
\newblock Unsupervised learning of object structure and dynamics from videos.
\newblock In {\em Advances in Neural Information Processing Systems}, pages
  92--102, 2019.

\bibitem{misra2016shuffle}
Ishan Misra, C~Lawrence Zitnick, and Martial Hebert.
\newblock Shuffle and learn: unsupervised learning using temporal order
  verification.
\newblock In {\em European Conference on Computer Vision}, pages 527--544.
  Springer, 2016.

\bibitem{oliu2018folded}
Marc Oliu, Javier Selva, and Sergio Escalera.
\newblock Folded recurrent neural networks for future video prediction.
\newblock In {\em Proceedings of the European Conference on Computer Vision
  (ECCV)}, pages 716--731, 2018.

\bibitem{oprea2020review}
Sergiu Oprea, Pablo Martinez-Gonzalez, Alberto Garcia-Garcia, John~Alejandro
  Castro-Vargas, Sergio Orts-Escolano, Jose Garcia-Rodriguez, and Antonis
  Argyros.
\newblock A review on deep learning techniques for video prediction.
\newblock {\em arXiv preprint arXiv:2004.05214}, 2020.

\bibitem{paszke2019pytorch}
Adam Paszke, Sam Gross, Francisco Massa, Adam Lerer, James Bradbury, Gregory
  Chanan, Trevor Killeen, Zeming Lin, Natalia Gimelshein, Luca Antiga, et~al.
\newblock Pytorch: An imperative style, high-performance deep learning library.
\newblock {\em arXiv preprint arXiv:1912.01703}, 2019.

\bibitem{ranzato2014video}
MarcAurelio Ranzato, Arthur Szlam, Joan Bruna, Michael Mathieu, Ronan
  Collobert, and Sumit Chopra.
\newblock Video (language) modeling: a baseline for generative models of
  natural videos.
\newblock {\em arXiv preprint arXiv:1412.6604}, 2014.

\bibitem{schuldt2004recognizing}
Christian Schuldt, Ivan Laptev, and Barbara Caputo.
\newblock Recognizing human actions: a local svm approach.
\newblock In {\em Proceedings of the 17th International Conference on Pattern
  Recognition, 2004. ICPR 2004.}, volume~3, pages 32--36. IEEE, 2004.

\bibitem{shi2015convolutional}
Xingjian Shi, Zhourong Chen, Hao Wang, Dit-Yan Yeung, Wai-Kin Wong, and
  Wang-chun Woo.
\newblock Convolutional lstm network: A machine learning approach for
  precipitation nowcasting.
\newblock {\em arXiv preprint arXiv:1506.04214}, 2015.

\bibitem{shi2017deep}
Xingjian Shi, Zhihan Gao, Leonard Lausen, Hao Wang, Dit-Yan Yeung, Wai-kin
  Wong, and Wang-chun Woo.
\newblock Deep learning for precipitation nowcasting: A benchmark and a new
  model.
\newblock In {\em Advances in neural information processing systems}, pages
  5617--5627, 2017.

\bibitem{srivastava2015unsupervised}
Nitish Srivastava, Elman Mansimov, and Ruslan Salakhudinov.
\newblock Unsupervised learning of video representations using lstms.
\newblock In {\em International conference on machine learning}, pages
  843--852, 2015.

\bibitem{tang2019pose}
Jilin Tang, Haoji Hu, Qiang Zhou, Hangguan Shan, Chuan Tian, and Tony~QS Quek.
\newblock Pose guided global and local gan for appearance preserving human
  video prediction.
\newblock In {\em 2019 IEEE International Conference on Image Processing
  (ICIP)}, pages 614--618. IEEE, 2019.

\bibitem{terwilliger2019recurrent}
Adam Terwilliger, Garrick Brazil, and Xiaoming Liu.
\newblock Recurrent flow-guided semantic forecasting.
\newblock In {\em 2019 IEEE Winter Conference on Applications of Computer
  Vision (WACV)}, pages 1703--1712. IEEE, 2019.

\bibitem{villegas2018hierarchical}
Ruben Villegas, Dumitru Erhan, Honglak Lee, et~al.
\newblock Hierarchical long-term video prediction without supervision.
\newblock In {\em International Conference on Machine Learning}, pages
  6038--6046. PMLR, 2018.

\bibitem{villegas2017decomposing}
Ruben Villegas, Jimei Yang, Seunghoon Hong, Xunyu Lin, and Honglak Lee.
\newblock Decomposing motion and content for natural video sequence prediction.
\newblock {\em arXiv preprint arXiv:1706.08033}, 2017.

\bibitem{wang2018predrnn++}
Yunbo Wang, Zhifeng Gao, Mingsheng Long, Jianmin Wang, and Philip~S Yu.
\newblock Predrnn++: Towards a resolution of the deep-in-time dilemma in
  spatiotemporal predictive learning.
\newblock {\em arXiv preprint arXiv:1804.06300}, 2018.

\bibitem{wang2018eidetic}
Yunbo Wang, Lu Jiang, Ming-Hsuan Yang, Li-Jia Li, Mingsheng Long, and Li
  Fei-Fei.
\newblock Eidetic 3d lstm: A model for video prediction and beyond.
\newblock In {\em International Conference on Learning Representations}, 2018.

\bibitem{wang2017predrnn}
Yunbo Wang, Mingsheng Long, Jianmin Wang, Zhifeng Gao, and S~Yu Philip.
\newblock Predrnn: Recurrent neural networks for predictive learning using
  spatiotemporal lstms.
\newblock In {\em Advances in Neural Information Processing Systems}, pages
  879--888, 2017.

\bibitem{wang2019memory}
Yunbo Wang, Jianjin Zhang, Hongyu Zhu, Mingsheng Long, Jianmin Wang, and
  Philip~S Yu.
\newblock Memory in memory: A predictive neural network for learning
  higher-order non-stationarity from spatiotemporal dynamics.
\newblock In {\em Proceedings of the IEEE Conference on Computer Vision and
  Pattern Recognition}, pages 9154--9162, 2019.

\bibitem{wang2004image}
Zhou Wang, Alan~C Bovik, Hamid~R Sheikh, and Eero~P Simoncelli.
\newblock Image quality assessment: from error visibility to structural
  similarity.
\newblock {\em IEEE transactions on image processing}, 13(4):600--612, 2004.

\bibitem{wu2021motionrnn}
Haixu Wu, Zhiyu Yao, Mingsheng Long, and Jianmin Wan.
\newblock Motionrnn: A flexible model for video prediction with
  spacetime-varying motions.
\newblock {\em arXiv preprint arXiv:2103.02243}, 2021.

\bibitem{wu2020future}
Yue Wu, Rongrong Gao, Jaesik Park, and Qifeng Chen.
\newblock Future video synthesis with object motion prediction.
\newblock In {\em Proceedings of the IEEE/CVF Conference on Computer Vision and
  Pattern Recognition}, pages 5539--5548, 2020.

\bibitem{xingjian2015convolutional}
SHI Xingjian, Zhourong Chen, Hao Wang, Dit-Yan Yeung, Wai-Kin Wong, and
  Wang-chun Woo.
\newblock Convolutional lstm network: A machine learning approach for
  precipitation nowcasting.
\newblock In {\em Advances in neural information processing systems}, pages
  802--810, 2015.

\bibitem{yang2018pose}
Ceyuan Yang, Zhe Wang, Xinge Zhu, Chen Huang, Jianping Shi, and Dahua Lin.
\newblock Pose guided human video generation.
\newblock In {\em Proceedings of the European Conference on Computer Vision
  (ECCV)}, pages 201--216, 2018.

\bibitem{yu2019crevnet}
Wei Yu, Yichao Lu, Steve Easterbrook, and Sanja Fidler.
\newblock Crevnet: Conditionally reversible video prediction.
\newblock {\em arXiv preprint arXiv:1910.11577}, 2019.

\bibitem{yu2019efficient}
Wei Yu, Yichao Lu, Steve Easterbrook, and Sanja Fidler.
\newblock Efficient and information-preserving future frame prediction and
  beyond.
\newblock In {\em International Conference on Learning Representations}, 2019.

\end{thebibliography}
}

\end{document}